\documentclass{article}

\usepackage{arxiv}

\usepackage{amsmath,amssymb,amsfonts}
\usepackage{algorithmic}
\usepackage{graphicx}
\usepackage{textcomp}
\usepackage{xcolor}

\usepackage[utf8]{inputenc}
\usepackage{amsfonts}
\usepackage{amsmath}
\usepackage{url}
\usepackage{multirow}

\usepackage[style=numeric,sorting=none]{biblatex}
\def\Bx{\mbox{\boldmath $x$}}
\def\By{\mbox{\boldmath $y$}}
\def\Bw{\mbox{\boldmath $w$}}
\def\Bu{\mbox{\boldmath $u$}}
\def\Bh{\mbox{\boldmath $h$}}

\def\Ba{\mbox{\boldmath $a$}}
\def\Br{\mbox{\boldmath $r$}}
\def\Bc{\mbox{\boldmath $c$}}
\def\Bz{\mbox{\boldmath $z$}}
\def\Bq{\mbox{\boldmath $q$}}
\def\Bo{\mbox{\boldmath $o$}}

\begin{document}

\title{Generate and Revise: \\ Reinforcement Learning in Neural Poetry}

\author{
  Andrea Zugarini\thanks{Corresponding author: http://sailab.diism.unisi.it/people/andrea-zugarini/} \\
  DINFO, DIISM\\
  University of Florence, University of Siena\\
  Florence 50139 Italy, Siena, 53100 Italy \\
  \texttt{andrea.zugarini@unifi.it}
  \And
  Luca Pasqualini\\
  DIISM\\
  University of Siena\\
  Siena 53100 Italy\\
  \texttt{pasqualini@diism.unisi.it}
   \And
  Stefano Melacci\\
  DIISM\\
  University of Siena\\
  Siena 53100 Italy\\
  \texttt{mela@diism.unisi.it}
   \And
  Marco Maggini\\
  DIISM\\
  University of Siena\\
  Siena 53100 Italy\\
  \texttt{maggini@diism.unisi.it}
}

\maketitle

\begin{abstract}
Writers, poets, singers usually do not create their compositions in just one breath.
Text is revisited, adjusted, modified, rephrased, even multiple times, in order to better convey meanings, emotions and feelings that the author wants to express. 
Amongst the noble written arts, Poetry is probably the one that needs to be elaborated the most, since the composition has to formally respect predefined meter and rhyming schemes.
In this paper, we propose a framework to generate poems that are repeatedly revisited and corrected, as humans do, in order to improve their overall quality.
We frame the problem of revising poems in the context of Reinforcement Learning and, in particular, using Proximal Policy Optimization. 
Our model generates poems from scratch and it learns to progressively adjust the generated text in order to match a target criterion. We evaluate this approach in the case of matching a rhyming scheme, without having any information on which words are responsible of creating rhymes and on how to coherently alter the poem words.
The proposed framework is general and, with an appropriate reward shaping, it can be applied to other text generation problems. 
\end{abstract}

\section{Introduction}\label{sec:info}
Developing machines that reproduce artistic behaviours and learn to be creative is a long-standing goal of the scientific community in the context of Artificial Intelligence \cite{BODEN1996267,colton2012computational}.
Recently, several researches focused on the case of the noble art of Poetry, motivated by success of Deep Learning approaches to Natural Language Processing (NLP) and, more specifically, to Natural Language Generation \cite{zhang2014chinese,wang2016chinese,yi2017generating,hopkins2017automatically,lau2018deep,zugarini2019neural}.
However, existing Machine Learning-based poem generators do not model the natural way poems are created by humans, i.e., poets usually do not create their compositions all in one breath.
Usually a poet revisits, rephrases, adjusts a poetry many times, before reaching a text that perfectly conveys their intended meanings and emotions.
In particular, a typical feature of poems is that the composition has also to formally respect predefined meter and rhyming schemes.

With the aim of developing an artificial agent that learns to mimic this behaviour, we design a framework to generate poems that are repeatedly revisited and corrected, in order to improve the overall quality of the poem. 
We frame this problem as a navigation task approached with Reinforcement Learning (RL), exploiting Proximal Policy Optimization (PPO) \cite{schulman2017proximal} that, to our best knowledge, is not commonly applied to Natural Language Generation, despite being an improved instance of the more common Vanilla Policy Gradient (VPG). In the task of generating and progressively editing the draft of a poem until it matches a target rhyming scheme, we show that PPO leads to better results than VPG. The agent is not informed about what a rhyme is and how to implement the considered scheme, making the task extremely challenging in an RL perspective. The agent generates a draft poem and it corrects the draft one word at a time. It not only understands that the ending words of each verse are the ones that are important with respect to the rhyming scheme, but that also other words of the poem might need to be adjusted to make the poem coherent with the rhyming words.
Despite the application to poetry generation, the proposed framework is general and it can be applied to other text generation problems, provided an opportune reward shaping. 

This paper is organized as follows. After discussing related work (Section~\ref{sec:related}), the neural models are described in Section~\ref{sec:framework}, while the RL-based poem revision dynamics is detailed in Section~\ref{sec:rl}. Experiments are reported in Section~\ref{sec:exp} and, finally, conclusions are drawn in Section~\ref{sec:concl}.

\subsection{Related Work}
\label{sec:related}
Early methods on Poetry Generation \cite{colton2012full} addressed the problem with rule-based techniques, whereas more recent approaches focused on learnable neural language models. 
The first deep learning solutions tackled Chinese Poetry. In \cite{zhang2014chinese}, authors combined convolutional and recurrent networks to generate quatrains. 
Afterwards, both \cite{yi2017generating} and \cite{wang2016chinese} proposed a sequence-to-sequence model with attention mechanisms.
In the context of English Poetry, transducers were exploited to generate poetic text \cite{hopkins2017automatically}. The generation structure (meter and rhyme) is learned from characters by cascading a module considering the context, with a weighted state transducer.
Recently, in Deep-speare \cite{lau2018deep}, the authors generated English quatrains with a combination of three neural models that share the same character-based embeddings. One network is a character-aware language model predicting at word level, another neural model learns the meter, and the last one identifies rhyming pairs. Generated quatrains are finally selected after a post-processing step from the output of the three modules.
In \cite{zugarini2019neural}, the authors focused on a single Italian poet, Dante Alighieri, by making use of a syllable-based language model, that was trained with a multi-stage procedure on non-poetic works of the same author and on a large Italian corpus.  

Reinforcement Learning has been recently used in several Natural Language Generation applications, such as Text Summarization \cite{paulus2017deep,chen2018fast,narayan2018ranking}, Machine Translation \cite{bentivogli2019machine} and Poem Generation \cite{yu2017seqgan,yi2018automatic} as well.
However, most of the proposed approaches exploit RL as a mean to make common evaluation metrics differentiable, such as BLEU and ROUGE scores \cite{belz2006comparing}. Of course, these metrics can be computed only in those tasks in which the target text (ground truth) is available. 
In \cite{yu2017seqgan} the authors extended Generative Adversarial Networks (GANs) \cite{goodfellow2014generative} to the generation of sequences of symbols, through Reinforcement Learning. The GAN discriminator is used as a reward signal for a RL-based language generator, and, among a variety of tasks, their framework was applied to Chinese quatrains generation.
In \cite{yi2018automatic}, a mutual Reinforcement Learning scheme was used to improve the quality of the generated Chinese quatrains.
In both works, different generic rewards were designed exploiting the simplest policy-based RL algorithm, i.e. Vanilla Policy Gradient. Surprisingly, Proximal Policy Optimization is less commonly used in the scope of Natural Language Generation, despite leading to a more robust and efficient RL algorithm \cite{tuan2018proximal}.

Our generate-and-revise framework is related to retrieve-and-edit seq2seq approaches \cite{guu2018generating,cao2018retrieve, li2018hybrid,weston2018retrieve,hossain2020simple}, where 
text generation reduces to an adaptation/paraphrasing of the retrieved template(s) related to the current input. The refinement process can be optimized with standard seq2seq learning algorithms because of the presence of revised targets. 
In our generate-and-revise instead, we neither start from retrieved templates, nor we have reference revisions. That is why we cast the problem as a navigation task and exploit RL to learn a revision policy that adjusts draft poems in order to improve their quality.

\section{Generate and Revise Poems}
\label{sec:framework}
Our framework is rooted on the idea that creating a poem is a multi-step process. First, the draft of a new poem is generated. Then, an iterative revision procedure is activated, in which the initial draft is progressively edited.
We model this problem by means of a \textit{generator}, that creates the draft, and a \textit{reviser}, that edits the draft up to the final version of the poem. The reviser is structured as an iterative procedure that, at each iteration, identifies a word of the poem which does not suit well the context in which it is located, and substitutes it with a better word.
At each step the reviser has to decide both \emph{which} word to replace and \emph{with what}. 
A straightforward approach to implement this idea is to design an RL agent that jointly addresses both the tasks. Thus, given an $m$-word poem with vocabulary size $|V|$, the agent has to choose among a large number of actions, i.e. $|V| \cdot m$, due to usually large $|V|$ (in the order of tens of thousands in our experiments). Therefore the problem quickly becomes extremely hard to tackle.

We keep the idea of exploiting an RL-based approach, but we decouple the problem implementing the reviser with two learnable models, namely the \textit{detector} and the \textit{prompter}, each of them responsible of one of the two aforementioned tasks, i.e., detecting a word to substitute (detector), and suggesting how to change a target word (prompter), respectively. The \textit{generator}, the \textit{detector}, and the \textit{prompter} are based on neural architectures, trained from scratch with appropriate criteria, while the \textit{detector} is fully developed by means of RL. The whole scheme is sketched in Fig.~\ref{fig:revise_framework_scketch}. The structure of this module allows us to reduce the action space of the RL procedure to (up to) $N$ words in the poem, making it independent on $|V|$. The prompter identifies the words in $V$ that are most compatible with the surrounding context.


In the following, the generator (Section~\ref{sec:CLM}), the detector (Section~\ref{sec:detector}) and the prompter (Section~\ref{sec:prompter_model}) will be described in detail, whereas the RL-dynamics of the detector are presented in Section~\ref{sec:rl}.

\subsection{Conditional Poem Generator}\label{sec:CLM}
The poem generation procedure is an instance of Natural Language Generation based on a learnable Language Model (LM). Before considering the specific details of Poetry, we describe the LM used in this work. Let us consider a sequence of tokens $(w_1, \ldots, w_{n+m})$ taken from a text corpus in a target language. For convenience in the description, let us divide the tokens into two sequences $\Bx$ and $\By$, where $\Bx= (x_1, \ldots, x_n)$ and $\By=(y_1, \ldots, y_m)$. The former ($\Bx$) is the context provided to the text generator from which to start the production of new text. More generally, $\Bx$ is a source of information that conditions the generation of $\By$ ($\Bx$ could also be empty).
The goal of the LM is to estimate the probability $p(\By)$, that is factorized as follows,
\begin{equation}
    p(\By) = \prod_{i=1}^{m} p(y_i|y_{<i},\Bx),
    \label{eq:nlg}
\end{equation}
being $y_{<i}$ a compact notation to indicate the words in the left context of $y_i$.
Notice that when the sequence $\Bx$ has size $n=0$, 
we fall back to the traditional LM formulation \cite{bengio2003neural}. The text generation is the outcome of sampling the next sequence $\By$ from (\ref{eq:nlg}).
Machine Translation, Text Summarization, Text Continuation, Poem Generation, and in general any sequence-to-sequence problem in NLP can be formulated as in (\ref{eq:nlg}).
The way $p(y_i|y_{<i},\Bx)$ will be related to the input sequence $\Bx$ depends on how strongly $\Bx$ is informative with respect to $\By$. 
 Problems in which the source sequence significantly biases the generation outcome are referred as \textit{non-open-ended} text generation, in contrast to \textit{open-ended} text generation, where the source sequence loosely correlates with the output $\By$ \cite{holtzman2019curious}. 

Poem Generation is an instance of \textit{open-ended} text generation. When starting to generate a novel poem from scratch, there is, of course, no source input sequence. After having generated a few verses or when starting from a few given verses, the next-verses generation can be conditioned using them (i.e., $\Bx$ contains previously given verses, while $\By$ is about the verses to be generated), but there is still a huge degree of freedom in the possible verses that can be generated, due to the intrinsically creative nature of Poetry. There might be several features to further constrain the LM with information that does not come from the input text, and, in this work, we consider two important features, that are the author $a$ and the target rhyme scheme $r$.
We update (\ref{eq:nlg}) by introducing the information on $a$ and $r$, 
\begin{equation}
    p(\By) = \prod_{i=1}^{m} p(y_i|y_{<i},\Bx, a, r).\label{eq:conditional_nlg}
\end{equation}
that is the reference equation on which our poem generation is based.

We model the distribution in Equation \ref{eq:conditional_nlg} by means of a sequence-to-sequence neural architecture with attention. 
Our LM is a variant of \cite{bahdanau2014neural}, similar to the one proposed in \cite{lau2018deep}, and it is based on an encoding-decoding scheme. The encoder is responsible of creating a compact representation of $\Bx$, while the decoder yields a probability distribution over the words in $V$ given the outcome of the encoder, and the conditioning signals $a$ and $r$ leading to $p(y_i|y_{<i},\Bx, a, r)$. 

\paragraph{Encoding.} The encoder of $\Bx$ computes a contextual representation of each word $x_j$ of the input sequence $\Bx$ ($n$ words), by means of a bidirectional LSTM (bi-LSTM). The output of this module is the set $H_{x}=\{\Bh_1, \ldots, \Bh_n \}$, being $\Bh_j$ the contextualized representation of the $j$-th word.
In detail, at each time step $j$, the bi-LSTM is fed with the concatenation of the word embedding $\Bw_j \in \mathbb{R}^d$ associated to $x_j$, and $\Bu_j \in \mathbb{R}^{r}$, a character-based representation of $x_j$. 
We indicate with $\overrightarrow{\Bh}_{j}$, $\overleftarrow{\Bh}_{j}$ the internal states of the bi-LSTM  processing the sequence of augmented word representations,
\begin{eqnarray*}
    \overrightarrow{\Bh}_{j} = \mathrm{\overrightarrow{LSTM}_{encx}}([\Bw_j, \Bu_j], \overrightarrow{\Bh}_{j-1}), \\
    \overleftarrow{\Bh}_{j} = \mathrm{\overleftarrow{LSTM}_{encx}}([\Bw_j, \Bu_j], \overleftarrow{\Bh}_{j+1}) ,
\end{eqnarray*}
where $\overrightarrow{LSTM}$, $\overleftarrow{LSTM}$ are the functions computed by the LSTMs in the two directions.
The final representation of the $j$-th word of the input sequence is $\Bh_j =[\overrightarrow{\Bh}_{j}, \overleftarrow{\Bh}_{j}]$. Overall, the encoder outputs $H_x=\{\Bh_1, \ldots, \Bh_n \}$.
The char-based representation $\Bu_j$ is obtained by processing the word characters with another bi-LSTM.
We augment $\Bw_j$ with a char-based representation to better encode sub-word information, that is crucial to capture rhyming schemes and meter in the poems.

\paragraph{Decoding.} 
The decoder is responsible of returning the distribution $p(y|y_{<i},\Bx,a,r)$ at each time index $i$, and, when used to generate text, to sample a word from $p$. 

We stack two recurrent layers. First an LSTM that computes at each time step $i$ a representation $\Bz_i$ given the previous word $y_{i-1}$ merged with author ($a$) and rhyme scheme ($r$) information encoded in form of embeddings $\Ba$ and $\Br$, obtaining:
\begin{equation*}
    \Bz_i = \mathrm{LSTM_{dec}}([\Bw_{i-1}, \Bu_{i-1}, \Ba, \Br] ,\Bz_{i-1}),
\end{equation*}
where $\Bw_{i-1}$ is the word embedding of $y_{i-1}$ and $\Bu_{i-1}$ is the character-aware word representation shared with the encoder.
Thanks to the inputs $\Ba$ and $\Br$, the state $\Bz_i$ includes author-specific and rhyme-scheme-specific information. 
This allows the system to generate text that is oriented toward the given author style and the target rhyme scheme.
The second recurrent layer is a Gated Recurrent Unit (GRU) cell \cite{chung2014empirical} that progressively fuses $\Bz_i$ with the context data in $H_{x}$, in order to create a further vector $\Bq_i \in \mathbb{R}^{d}$ that compactly includes all the conditioning signals of (\ref{eq:conditional_nlg}). First, an attention mechanism \cite{bahdanau2014neural} is applied over the encoding of the words of $\Bx$, i.e, on their contextualized representations collected in $H_{x}$, yielding an attention-based representation  $\Bc_i$ of $\Bx$,
\begin{equation*}
    \Bc_i = \mathrm{attn}(\Bz_i, H_{x}).
\end{equation*}
Then, the concatenation of $\Bc_i$ with the representation $\Bz_i$ of the triple $(y_{<i},a,r)$ is processed by a GRU cell, 
\begin{equation*}
    \Bq_{i} = \mathrm{GRU}([\Bc_i, \Bz_i], \Bz_{i-1}),
\end{equation*}
Finally, the distribution $p(y|y_{<i},\Bx,a,r)$ is obtained through a linear projection of $\Bq_i$ with the transposed embedding matrix $E' \in \mathbb{R}^{d \times |V|}$, and then applying the softmax function. 
The model is trained to maximize $p(\By)$ on a text corpora of poems (see Section~\ref{sec:exp}).

\paragraph{Generation.} Poems are generated sampling from $p$. As a matter of fact, the sampling strategy plays a crucial role in the quality of the generated text, and it has been recently shown to have a major impact in Natural Language Generation \cite{see2019massively}.  We preferred nucleus (top-$p$) sampling, with $p=0.9$, to generate quatrains over multinomial and top-$k$ sampling. 
We indicate with $\Bo$ the sequence of words sampled from $p$ that will consitute a draft poem.
The drafts poems generated by the model will be then revised by the joint work of detector and prompter modules. 

\subsection{Detector}
\label{sec:detector}
Once we have generated a draft poem using the model of Section~\ref{sec:CLM}, a detection module learns to select the next word of the draft that needs to be revised.
The detector is a neural model that yields a probability distribution $\pi(o_i | \Bo, a, r)$ over the $N$ words of the poem. Of course, in order to detect which words to replace, it is important to take into account the author and rhyme information.
In detail, the words of poem $\Bo$ are encoded by a network that is analogous to the encoder of $\Bx$ in Section~\ref{sec:CLM}. The word representations, collected in $H_{o}$, are processed by an attention mechanisms $\mathrm{attn}_{det}$, building a compact embedding of the whole poem that is also function of the author and of the rhyme scheme.
Then, a Multi-Layer Perceptron (MLP) with softmax activation in the output layer returns the probability over the $N$ words,
\begin{equation}
\pi(o_j | \Bo, a, r) = \mathrm{MLP}_j(\mathrm{attn}_{det}([\Ba,\Br], H_{o}))
\label{eq:detect2}
\end{equation}
being $\mathrm{MLP}_j$ the $j$-th output unit.
Multinomial sampling applied to $\pi$ leads to the selection of the word(s) that should be replaced. This module is trained by RL, as we will describe in Section~\ref{sec:rl}.

\subsection{Prompter}
\label{sec:prompter_model}
The role of the prompter module is to provide valid candidates to replace the word previously selected by the detector of Section~\ref{sec:detector}.
The prompter module solves the problem of modeling language given the left-right contexts of each word, that can be formulated following an approach similar to the one exploited by the conditional LM of (\ref{eq:conditional_nlg}). 
Thus, given an author $a$ and a rhyme scheme $r$, we use a neural model to learn the following distribution from data,
\begin{equation}
        p(\Bo) = \prod_{i=1}^{N} p(o_i|o_{<i}, o_{>i}, a, r),
        \label{eq:bilm}
\end{equation}
being $o_{<i}$, $o_{>i}$ the words in left and right context of $o_i$, respectively. Once p(\Bo) has been learnt, we can sample $p(o_i|o_{<i}, o_{>i}, a, r)$ to get one or more candidate words for replacing the selected one.

The prompter network follows the context encoding schemes of  \cite{melamud2016context2vec} and  \cite{marra2018unsupervised}. 
In particular, the words of poem $\Bo$ are encoded by a network that computes representations of the left and right contexts around each target word, discarding the target word itself.
Differently from the encoding of $\Bo$ in Section~\ref{sec:detector}, here the final representation of the $j$-th word is then $[\overrightarrow{\Bh}_{j-1},\overleftarrow{\Bh}_{j+1}]$.\footnote{In our implementation, we used the same LSTMs when encoding data in the detector and in the prompter module.}
This representation is concatenated with the author embedding $\Ba$ and the rhyme scheme embedding $\Br$, followed by a learnable linear layer with softmax activation that projects the concatenated vector to the space of vocabulary indices. Including $\Ba$ and $\Br$ in the prompter module is crucial in order to allow the network to learn how to revise a target word  in function of the poet and rhyme scheme.
Candidate(s) for replacing the selected word are sampled from $p(o_i|o_{<i}, o_{>i}, a, r)$, as discussed in the Poem Generator of Section~\ref{sec:CLM}. In this case we used top-$k$ sampling ($k=50$) to have a large pool of candidates. The prompter is trained to maximize $p(\Bo)$ on a text corpora of poems (Section~\ref{sec:exp}).

\begin{figure*}[htbp]
\centering
\includegraphics[width=0.8\textwidth]{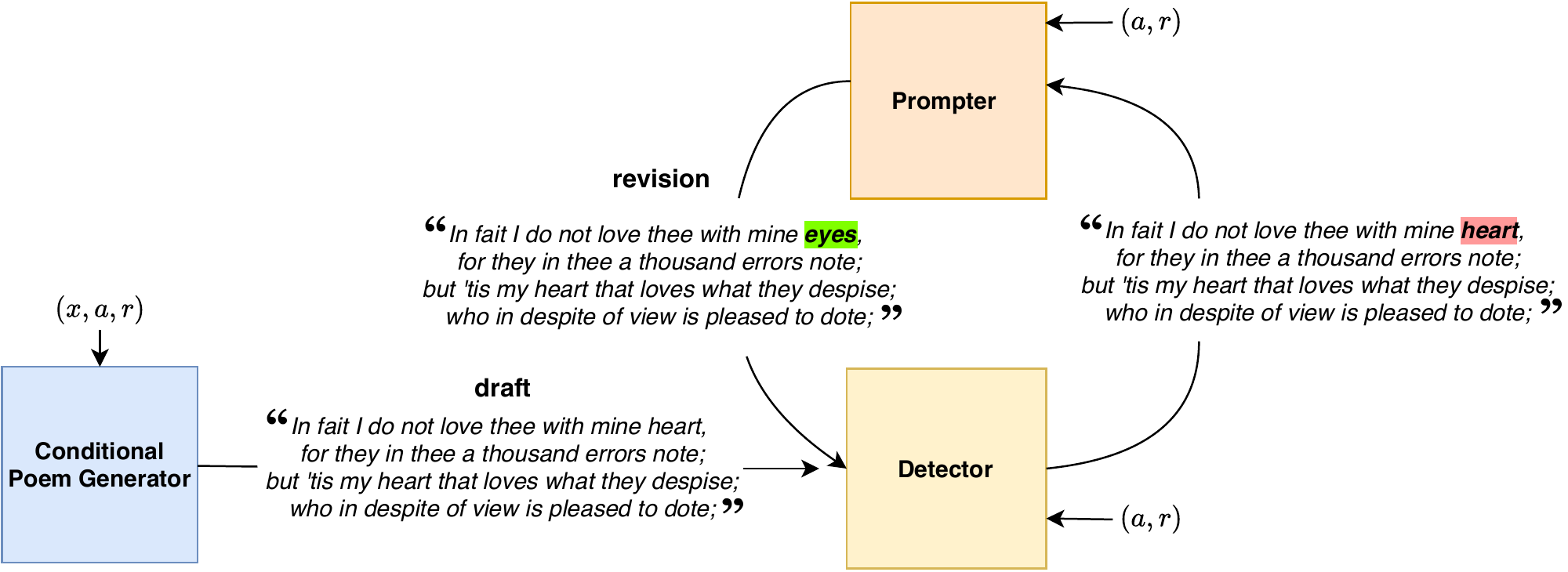}
\caption{Overall Generate and Revise scheme on an example poem. The conditional poem generator (light blue module) produces a draft poem, which is iteratively revised by the detector (pale yellow) - Prompter (light orange) modules until satisfaction of a certain criteria. At each step the detector identifies the word to replace, \textit{heart} highlighted in red, while the prompter is responsible for finding the substitute, \textit{eyes} highlighted in green.}
	\label{fig:revise_framework_scketch}
\end{figure*}

\section{Revision as a Navigation Task}
\label{sec:rl}

Once the poem generator and the prompter modules have been trained, the task of revising a generated poem consists in detecting which words to change and letting the prompter replace them. If we assume to change one word at a time, we can easily consider this task as a decision process in the space of the dictionary words $V$. Each decision defines which word to change at that a given step, and the prompter replaces it with a suitable candidate. The sequence of decisions is the \textit{policy} of an agent whose goal is to improve the text, according to a given reward function. Text revision stops when a satisfying score has been reached.
This task may be cast as a navigation problem, where the current \textit{state} of the agent is identified by the sequence of the words in the current text revision. This allows us to reformulate the problem as an RL task where the navigation space is the environment \cite{babu2016autonomous}, while the decisions are identified by actions executed by the agent in the environment. We provide a brief introduction of Reinforcement Learning in Appendix~\ref{app:rl_intro}.

A RL task can be framed as a sequential decision-making problem in which, at each step $t$, the agent observes a state $S_t\in\mathcal{S}$ from the environment, and then selects an action $A_t\in\mathcal{A}$. The environment yields a numerical reward $R_{t+1}\in\mathcal{R}$ and then it moves to the next state $S_{t+1}$. This interaction gives raise to a \emph{trajectory} of random variables. In our task, since words are elements of the vocabulary $V$, we have that $\mathcal{S}$ is the space of the poems of length $N$ with words from $V$ for the target author $a$ and with rhyming scheme $r$, $\mathcal{A}$ is the set of indices of the word positions in the poem plus the do-nothing action, while $\mathcal{R}\subset\mathbb{R}$.
To define a reward function we use the shortest path problem formulation. The agent aims at reaching the final text revision in the least amount of steps. Conventionally this means that the reward $R_t$ is defined as a negative number for each state not at the goal state position and a positive number or zero when the goal state is reached. Formally, if $\Bo_{t}$ is the poem revision at step $t$, we have
\begin{eqnarray}
    A_t &=& \hat{A}_t \in \bigcup_{g=1}^{N+1}\{g\}   \\
    \label{eq:s} S_{t+1} &=& ( \Bo_{t+1}, a, r ) \\
    R_{t+1} &=&
    \begin{cases}
        1              & \text{ if } S_{t+1} = S_f \\
        -1               & \text{ otherwise}
    \end{cases}\label{eq:reward}
\end{eqnarray}
where $S_f$ is the goal state in which the text is not revised anymore.

The natural connection between the modules presented in Section~\ref{sec:framework} and the RL-based setting is easily established once we redefine (\ref{eq:detect2}) as the probability of an action in the state described by the triple $(\Bo,a,r)$, that perfectly suits the definition in (\ref{eq:s}), yielding a \textit{policy} function. Using Deep Neural Networks (DNNs) to approximate the RL-related functions, as we do in the case of the probability distribution over the action space $\pi$, is a pretty common approach in nowadays RL-based problems (see, e.g., \cite{babu2016autonomous}). In the following descriptions, we compactly rewrite (\ref{eq:detect2}) adding the symbol $\theta$ to refer to the network weights that are learned by means of the RL procedure, i.e., $\pi(\cdot|\cdot;\theta)$. 
Policy Gradient methods are suitable for navigation tasks, as shown in \cite{knudson2010policy}, especially when the states' space becomes large \cite{pasqualini2020pseudo}. In such spaces often off-policy algorithms (like Q-Learning) are indeed observed to be unable to converge.
In this work, we compare two on-policy RL algorithms: Vanilla Policy Gradient (Section~\ref{sec:vpg}) and Proximal Policy Optimization (Section~\ref{sec:ppo}). 

\subsection{Vanilla Policy Gradient}
\label{sec:vpg}
Vanilla Policy Gradient (VPG) \cite{sutton2000policy} is an on-policy RL algorithm whose aim is to learn a policy without using $q$-values as a proxy. This is obtained increasing the probabilities of actions that lead to higher return, and decreasing the probabilities of actions that lead to lower return. Actions are usually sampled from a multinomial distribution for discrete actions' spaces and from a normal distribution for continuous action spaces.
VPG works by updating policy parameters $\theta$ via stochastic gradient ascent on policy performance over a buffer built from a certain number of trajectories,
\[
\theta \longleftarrow \theta + \alpha \nabla_{\theta} J(\pi(\cdot|\cdot; \theta))
\] 
where $J(\pi(\cdot|\cdot; \theta))$ denotes the expected finite-horizon undiscounted return of the policy and $\nabla_{\theta}$ its gradient with respect to $\theta$ ($\alpha > 0$). In order to compute $J(\pi(\cdot|\cdot; \theta))$, the algorithm requires to evaluate further actions for each state $s$ in the buffer. In this paper, we use Generalized Advantage Estimation (GAE) \cite{schulman2015high} to compute such actions, and the obtained rewards are saved and normalized with respect to 
``when'' they are collected (the so called rewards-to-go). These are solutions reported in literature to be stable and to improve overall training performance of the model.

\subsection{Proximal Policy Optimization}
\label{sec:ppo}
Proximal Policy Optimization (PPO) \cite{schulman2017proximal} is another on-policy RL algorithm which improves upon VPG. It is considered the state-of-the-art in policy optimization methods and it is a modified version of Trust Region Policy Optimization (TRPO) \cite{schulman2015trust}. Both methods try to take the biggest possible improvement step on a policy using the currently available data, without stepping ``too far'' and making the performance collapse. This is done by maximizing a surrogate objective, subject to a constraint on policy update quantity, where such constraint depends on the KL-divergence between the old policy and the new policy after the update. Specifically, PPO  uses  a  clipped  objective to  heuristically constrain the KL-divergence,
\[
\max_\theta{ \mathbb{E}[\min(\rho_t \bar{A_t}, \textrm{clip}(\rho_t, 1 - \epsilon, 1 + \epsilon) \cdot \bar{A_t})}, 
\]
where $\rho_t = \frac{\pi(A_t | S_t; \theta_t)}{\pi_{\theta_{old}}(A_t | S_t; \theta_{t-1})}$ is a policy ratio, $\textrm{clip}(\rho_t,\cdot,\cdot)$ clips $\rho_t$  in the interval defined by the last two arguments, $\epsilon$ is an hyperparameter (we set $\epsilon = 0.2$), $\bar{A_t}$ is the estimated advantage function at time step $t$ and $A_t$, $S_t$ are respectively the action and the state at time step $t$.
In the implementation used in this paper, when parameters $\theta$ are updated over a buffer of trajectories, the update process is early stopped if the constraint is not respected, thus avoiding the new policy to step ``too far'' from the previous one. 


\section{Experiments}
\label{sec:exp}
We collected poems in English language from the Project Gutenberg\footnote{\url{https://www.gutenberg.org/}} using the GutenTag tool \cite{brooke2015gutentag} to filter out non-poetic work and collections. We also discarded non-English contents that occasionally appeared in the retrieved documents.
Poems are organized in stanzas, according to their XML-based description. 
Each stanza was then divided into quatrains, if not already in such format, and we assigned a rhyming scheme to each stanza, from a fixed dictionary of rhyming schemes. Rhymes were automatically detected with the Pronouncing library\footnote{\url{https://pypi.org/project/pronouncing/}} and a few additional heuristic rules to cover most of the undetected rhymes. Long poems without any rhyming pattern were discarded as well.
We used the meta-information about the author to define the authorship of the stanza, when available. We considered the most $768$ frequent authors, the rest was marked as unknown.
Overall, we obtained $757,891$ quatrains, divided in three sets of the sizes $684,100$, $36,006$ and $37,785$, respectively, used to train, validate and test the models. 
We limited the word vocabulary to the most frequent $50,000$ words, assigning an embedding of size $300$ for all the models in all the experiments. The maximum sequence length of a quatrain has been set to $50$, longer verses were truncated.

We define multiple experimental settings and tasks in order to evaluate the quality of the each module proposed in this work, up to the entire system that includes all the modules and the full pipeline of generation and iterative revision.


\subsection{Conditional Poem Generator}
\label{subsec:exp_clm}
While the proposed generator of Section~\ref{sec:CLM}, follows an established neural architecture, the innovative elements we introduce in this work are about the poem-related conditional features, \emph{author} and  \emph{rhyme scheme}, and their use in Poetry with character-aware representations.
We considered the task of generating a quatrain $\By$ given the context sequence $\Bx$ that is the previous quatrain, where the rhyme scheme is a symbol indicating the rhymes of an eight-verse poem. We considered the $50$ most frequent rhyme schemes of size eight. 
The architecture hyper-parameters were commonly selected by choosing the best configuration on the validation set for the vanilla (i.e., not conditioned by author and rhyme) generator. The bi-LSTM state encoding the context sequence $\Bx$ was set to $512$, as the state of the decoder $LSTM_{enct}$, and the GRU cell as well. Author and rhyme embedding sizes were set to $128$ and $256$, respectively.

We compared in terms of generation perplexity a model trained with or without any of the newly introduced conditional features, reporting results in Table~\ref{tab:clm_ppl}.
The conditional features allows the LM to be more accurate, that is an important result considering the open-ended challenging nature of the poem generation task.

\subsection{Prompter}
\label{subsec:exp_prompter}
A similar analysis was followed to evaluate the quality of the prompter model of Section~\ref{sec:prompter_model}. 
In particular, we trained a prompter model on single quatrains, enforcing it to learn how to predict a word given its context. We used a bi-LSTM state of $1024$ units.
Again, the role of the new conditional features is what we are mostly interested in and, observing results of Table~\ref{tab:pro}, we can see that they improve the suggestion quality. This result is in line with the case of Section~\ref{subsec:exp_clm}, confirming the importance of further poem-related information.

\begin{table}[htbp]
\caption{Perplexity measured on the validation (Val) and test (Test) sets of the poem generator, trained with or without conditional features.}
\centering
\begin{tabular}{lll}
 & Val & Test \\
 \hline
 Vanilla Generator & 52.98 & 59.78\\
  Conditional Generator& \textbf{51.40} & \textbf{54.86}\\ 
 \hline
\end{tabular}
\label{tab:clm_ppl}
\end{table}

\begin{table}[htbp]
\caption{Perplexity measured on the validation (Val) and test (Test) sets of the prompter module, trained with or without conditional features.}
\centering
\begin{tabular}{lll}
 & Val & Test \\
 \hline
  Vanilla Prompter & 14.09 & 14.78 \\
  Conditional Prompter & \textbf{12.90} & \textbf{13.40} \\ 
 \hline
\end{tabular}
\label{tab:pro}
\end{table}

\subsection{Revision as a Navigation Task}
\label{subsec:detector_exp}
In order to show the quality of the detector module and that approaching text correction as shortest path problem is feasible, we created ``corrupted'' poems from real poems in the dataset by replacing one or more words in random positions with words sampled from the entire vocabulary $V$.\footnote{Frequent words are sampled as replacement more often than rare ones.}
The agent operates in an environment where each episode starts with a corrupted poem, and it has to learn to reconstruct the original not-corrupted poem, selecting at each step which word to change. 
In this artificial setting we assume that, once the agent picks which word to substitute, a perfect prompter (oracle) will replace it with the ground truth, i.e. the word originally positioned there in the real poem. 
This means that after each agent action, the selected position will be either replaced with the original word, in case of a corrupted word, or nothing will be changed, in case of a correct word. 
The navigation terminates when the goal state is reached, that occurs after all the corrupted words are removed from the poem. 

The MLP predicting actions has a single hidden layer of size $512$. 
We performed different experiments over this poem reconstruction environment, using a PPO-based agent. Each experiment differs in the number of poems that the agent has to fix, and the number of words perturbed in the poem.
We considered \{1, 10, 100\} poems that, at the beginning of each episode, are randomly  ``corrupted'' by altering $1$ or $3$ (referred to as ``multiple'') words in the original poem.
Please note that even in the simplest case, the experiment with one poem only and a single perturbed word, the number of generated ``corrupted'' poems is huge, $|V|^{|x|}$, where $|x|$ is the poem length. 

The PPO-agent is trained for $10$ volleys in each experiment, with $1,000$/$20,000$/$200,000$ episodes in the experiments with $1$/$10$/$100$ poems, respectively. We set the maximum episode length to $10$ steps. Hence, the reward varies between $[-10,1]$ where $1$ corresponds to the case in which the agent immediately identifies the ``corrupted'' word (when there is only 1 corrupted word), and $-10$ indicates a full failure.
Results are shown in Table \ref{tab:baseline_ppo_single}. The ``Volley 0'' column defines the average total reward at the end of the first volley, while the ``Volley 9'' column defines such value at the end of the last volley. 
\begin{table}[htbp]
\caption{Results of the experiments with the PPO-based agent on poem reconstruction task of Section~\ref{subsec:detector_exp}. The averaged total rewards after the first volley and the last volley are reported, respectively.}
\centering
\begin{tabular}{lcc}
 & $R$ Volley $0$ & $R$ Volley $9$ \\
 \hline
  1 Poem & -7.866 & -0.425 \\
  1 Poem (multiple) & -9.092 & -3.126 \\
  10 Poems & -7.793 & 0.293 \\
  100 Poems & -8.110 & -6.389 \\
 \hline
\end{tabular}
\label{tab:baseline_ppo_single}
\end{table}
The reward value improves during the training volleys, while increasing the number of poems makes the problem exponentially more complex. Due to the dynamic perturbation of the poems on each volley, the policy learned by the agent is not tied to a set of perturbed words, but it is general over the set of poems the agent is trained on.
However, independently from the amount of poems, results confirm that the revision problem can be framed into a shortest path problem and addressed by RL using PPO, in a varying amount of time.

\begin{table*}[htbp]
\caption{Two examples of generated poems with generate and revise approach given a target rhyme scheme, before and after the revision iterative steps.}
\centering
\footnotesize
\begin{tabular}{ccc}
Rhyme scheme & Draft & Revision \\ \hline
\multirow{4}{*}{AABB} & \textit{the mist that made us sweat and \textbf{ache}} & \textit{the mist that made us sweat and \textbf{chill}}\\ 
& \textit{with toil, from doing good or ill,} &\textit{with toil, from doing good or ill,} \\
& \textit{the hour when we were led to play} & \textit{the hour when we were led to play}\\
& \textit{the children of the people's \textbf{brood},} &\textit{the children of the people's \textbf{way},} \\
& & \\
\multirow{4}{*}{ABBB} &\textit{and when, above, the winter's snow} & \textit{and when, above, the winter's snow} \\
& \textit{has risen in the wintry \textbf{sky}} & \textit{has risen in the wintry night \textbf{away}}\\
& \textit{and leaves their path to cloud's decay,} & \textit{and leaves their path to cloud's decay,}\\
&\textit{and life is spent, and life is drear\textbf{,}} & \textit{and life is spent, and life is drear \textbf{today}}\\
\hline
\end{tabular}
\label{tab:poem_examples}
\end{table*}

\subsection{Generate and Revise Poems}
\label{subsect:all}
Now we consider the complete system in which all the modules are active as in Fig.~\ref{fig:revise_framework_scketch}. We focused on the task of generating poems and progressively revising them, in which the agent goal is to substitute words so that the poem matches a target rhyme scheme. 
Episodes begin with poems generated by the conditional generator. This task is significantly more challenging than the previously described ones, since there is no ground truth for generated poems, and words replacements are provided by the prompter model described in Section \ref{sec:prompter_model}.
Therefore, we let the model free to change any word in the quatrain, without restricting the agent actions to words at the end of each verse. Basically, the agent does not know that rhymes are related to the ending words of some verses, while the only information it receives is the reward (or penalty) signal that tells if the poem fulfills the target rhyming scheme or not.

We ran several experiments comparing PPO with VPG, varying in each experiment the number of poems to revise in the environment in $\{10$, $100$, $200$, $500$, $1,000\}$. We set to the number of training steps per volley at $10,000$ for the experiment with $10$ poems, and we increase it to $100,000$ in the other experiments. Additionally, we considered another experiment, indicated as \texttt{dynamic}, in the most difficult scenario, i.e., where the environment spawns new, unseen, artificially generated quatrains at each episode. In such a case we report results of PPO only, because using VPG always resulted in a failure.
An episode ends either when the target rhyme scheme is matched or after $30$ steps, that corresponds to the maximum episode length. Therefore, the reward of an episode ranges in the interval $[-30, 1]$. 
Differently from our previous work \cite{zugarini2019neural}, we do not carry out human evaluations, since rhyme matching can be quantitatively measured through the reward. Indeed, the reward is a direct way to assess the revised poem quality, because it is proportional to the number of steps needed for adjusting the target rhyme scheme. In particular, from Equation~\ref{eq:reward} we can observe that the number of revising steps in an episode (i.e. where reaching the goal state $S_f$) is equivalent to $|R_f| + 2$.

\begin{figure}[htbp]
    \centering
    \includegraphics[width=0.45\textwidth]{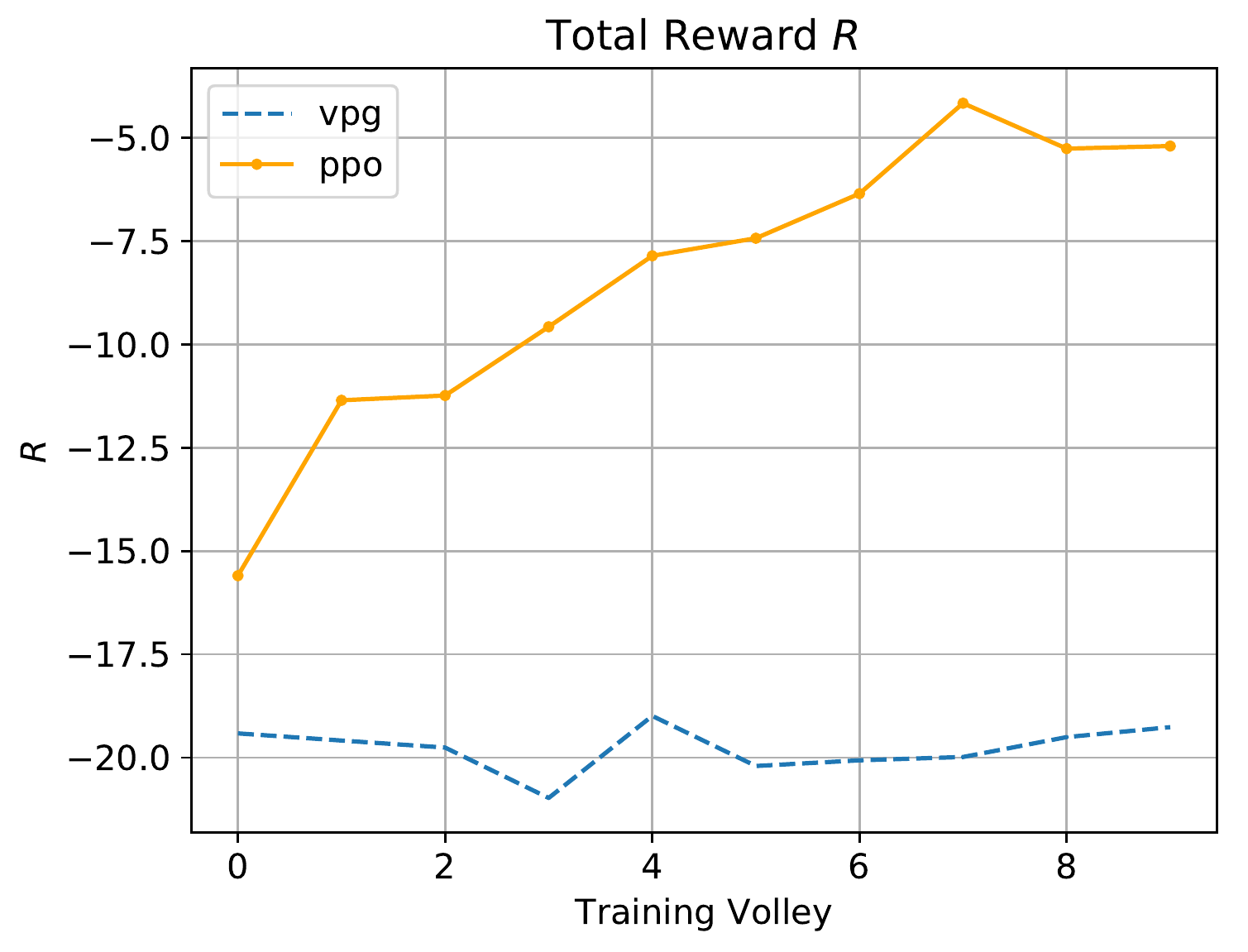}
    \vskip -3mm
    \caption{Rewards yielded by using PPO and VPG with respect to the number training volleys, in the experiment of Section \ref{subsect:all} with $100$ poems in the environment.}
    \label{fig:rhyme_exp_ppo_vpg_100}
\end{figure}

Results are presented in Table \ref{tab:exp_rhyme}. 
We can see that, while the agent improves the reward in all the experiments with PPO, learning with VPG is not stable, and performs poorly. The superiority of PPO over VPG is also illustrated in Fig. \ref{fig:rhyme_exp_ppo_vpg_100}, where we can see the instability of VPG in contrast to the steady progresses of PPO. Even if the task is very challenging, the model is able to strongly improve the average $R$ score, thus indicating that it is actually moving the right steps in progressively fixing the rhymes. We also report in Table~\ref{tab:poem_examples} two examples of draft revisions obtained with the agent trained with PPO in the \texttt{dynamic} environment. 

\begin{table}[htbp]
\caption{VPG vs PPO: Reward on the experiment of Section~\ref{subsect:all} with 10, 100, 200, 500 and 1000 poems. PPO is also evaluated with an environment that continuously generates new drafts (\texttt{dynamic}).}
\centering
\begin{tabular}{lccc}
 &    N poems   &  $R$ first Volley  &   $R$ last Volley\\ \hline
VPG&\multirow{2}{*}{10}  & -18.752   &  -14.630\\
PPO                  &  & -10.239 & \textbf{-1.186} \\ \hline
VPG &\multirow{2}{*}{100}  & -19.415 & -19.264 \\
PPO                  &  & -15.598 &   \textbf{-5.200}\\ \hline
VPG&\multirow{2}{*}{200}  & -20.950 & -18.432 \\
PPO                  &  & -15.323   & \textbf{-3.757}\\ \hline
VPG&\multirow{2}{*}{500}  & -21.191 & -19.623 \\
PPO                  & & -15.043  & \textbf{-7.780} \\ \hline
VPG&\multirow{2}{*}{1,000}  & -26.150 & -21.179 \\
PPO                  &  & -11.579 & \textbf{-9.733}\\ \hline
PPO & \texttt{dynamic} & -14.796 & -12.415\\ \hline
\end{tabular}\label{tab:exp_rhyme}
\end{table}

\section{Conclusions and Future Work}
\label{sec:concl}
In this paper we presented an innovative way of implementing the notion of creativity in a machine. Considering the task of automatically generating new poems, we proposed a model that implements the human-like behaviour of writing a draft and revising it multiple times. We proposed to create drafts that are conditioned to author and rhyme information, while the revision process is built around an iterative procedure that can be described as a navigation problem and solved with Reinforcement Learning with Proximal Policy Optimization, that significantly outperformed Vanilla Policy Gradient.  Multiple experiments confirmed that the proposed approach is feasible and that it allows the machine to learn how to revise text, even if it is not explicitly instructed on which portion of text it should revise. The proposed framework is also general enough to be eventually applied to other text generation tasks, that is what we are going to do in future work.

\printbibliography

\appendix
\section{Reinforcement Learning}
\label{app:rl_intro}
Reinforcement Learning\footnote{For a comprehensive introduction to RL, see sections from $1.1$ to $1.6$ in \cite{sutton2018reinforcement}.} (RL) is learning what to do in order to accumulate as much reward as possible during the course of actions. This very general description, known as \emph{the RL problem}, can be framed as a sequential decision-making problem as follows.

Let us consider an \emph{agent} interacting with an \emph{environment}, through a set of possible \emph{actions} that depends on the current situation, namely the \emph{state}.
An action affects the environment, therefore after each action the state will change. Some states are ``better'' than others, and the goodness of the state can be numerically quantifies with a value, called \emph{reward}. The pair ``state, reward'' may possibly be drawn from a joint probability distribution, called the \emph{model} or the \emph{dynamics} of the environment. The agent will choose actions according to a certain strategy, called \emph{policy} in the RL setting. The RL problem can then be stated as finding a policy maximizing the expected value of the total reward accumulated during the interaction agent-environment.

The RL problem implicitly assumes that the joint probability distribution of $S_{t+1},R_{t+1}$, i.e. the state of the environment and the reward obtained at next time step $t+1$ depend only on the past via $S_t$ and $A_t$, corresponding to the state of the environment and the action executed by the agent at time step $t$. In fact, the environment is fed only with the last action, and no other data from the history. This means that, for a fixed policy, the corresponding stochastic process $\{S_{t}\}$ is Markovian. 
When the agent experiences a trajectory, or episode, starting at time $t$, it accumulates a \emph{discounted return} $G_t$:
\[
G_t:=R_{t+1}+\gamma R_{t+2}+\gamma^2 R_{t+3}+\dots=\sum_{k=0}^\infty\gamma^k R_{t+k+1},\qquad\gamma\in[0,1].
\]
The return $G_t$ is a random variable, whose probability distribution depends not only on the environment dynamics, but also on how the agent chooses actions in a certain state $s$. Choices of actions are encoded by the policy, i.e.\ a discrete probability distribution $\pi$ on $\mathcal{A}$:
\begin{equation*}\label{eq:policy}
\pi(a|s):=\pi(a,s):=\text{Pr}(A_t=a|S_t=s).
\end{equation*}
A discount factor $\gamma<1$ is used mainly when rewards far in the future are less and less reliable or important, or in \emph{continuing} tasks, that is, when the trajectories do not decompose naturally into \emph{episodes}. 

The average return from a state $s$, that is, the average total reward the agent can accumulate starting from $s$, represents how good is the state $s$ for the agent \emph{following the policy $\pi$}, and it is called \emph{state-value} function:
\[
v_{\pi}(s):=\\E_\pi[G_t|S_t=s].
\]
Likewise, one can define the \emph{action-value} function (known also as \emph{quality} or \emph{q-value}), encoding how good is \emph{choosing an action $a$ from $s$ and then following the policy $\pi$}:
\[
q_{\pi}(s,a):=\\E_\pi[G_t|S_t=s,A_t=a].
\]

In most problems, like the one at hand, we have only a \emph{partial knowledge of the environment dynamics}. This can be overcome by \emph{sampling trajectories} $S_t=s,A_t=a,R_{t+1},S_{t+1},A_{t+1},R_{t+2},\dots$. Policy Gradient (PG) algorithms estimate directly the policy $\pi(a|s;\theta)$ from sampled trajectories, without using a value function. The parameters vector $\theta_t$ at time $t$ is often approximated by a neural network in a Deep Reinforcement Learning (DRL) fashion and it is modified to maximize a suitable scalar performance function $J(\theta)$, with the gradient ascent update rule:
\[
\theta_{t+1}:=\theta_t+\alpha\widehat{\nabla J(\theta_t)}.
\]
Here the \emph{learning rate} $\alpha$ is the step size of the gradient ascent algorithm, determining how much we are trying to improve the policy at each update, and $\widehat{\nabla J(\theta_t)}$ is any estimate of the performance gradient $\nabla J(\theta)$ of the policy. Different choices for the estimator corresponds to different PG algorithms. In this paper we use two PG algorithms, Vanilla Policy Gradient (Section~\ref{sec:vpg}) and Proximal Policy Optimization (Section~\ref{sec:ppo}).

\end{document}